\documentclass[10pt,twocolumn,letterpaper]{article}

\usepackage{cvpr}
\usepackage{xcolor}
\usepackage{times}
\usepackage{epsfig}
\usepackage{graphicx}
\usepackage{amsmath}
\usepackage{amssymb}
\usepackage{adjustbox}
\usepackage{tabularx}
\usepackage{algorithm}
\usepackage{algpseudocode}
\usepackage{multirow}
\usepackage{subcaption}

\graphicspath{{data/}}
\newcolumntype{s}{D{.}{.}{1.2}}
\newcolumntype{d}{D{.}{.}{2.1}}

\newcolumntype{A}{>{\centering\arraybackslash}X}

\graphicspath{{data/}}
\ifdefined\MYFLAG

\else

\fi

\newcolumntype{b}{>{\hsize=2.3\hsize}X}
\newcolumntype{s}{>{\hsize=.45\hsize}X}
\newcolumntype{m}{>{\hsize=.9\hsize}X}

\usepackage[pagebackref=true,breaklinks=true,letterpaper=true,colorlinks,bookmarks=false]{hyperref}

\cvprfinalcopy 

\def\cvprPaperID{622} 

\ifcvprfinal\fi
\begin{document}

\title{Scene Graph Generation with External Knowledge and Image Reconstruction}

\author{Jiuxiang Gu$^{1}$\thanks{This work was done during the author's internship at Adobe Research.}, Handong Zhao$^2$, Zhe Lin$^2$, Sheng Li$^3$, Jianfei Cai$^1$, Mingyang Ling$^4$\\
	$^1$ ROSE Lab, Interdisciplinary Graduate School, Nanyang Technological University, Singapore \\
	$^2$ Adobe Research, USA $^3$ University of Georgia, USA $^4$ Google Cloud AI, USA \\
    \tt\small \{jgu004, asjfcai\}@ntu.edu.sg, \{hazhao, zlin\}@adobe.com\\ 
    \tt\small sheng.li@uga.edu, mingyangling@google.com
}

\maketitle
\thispagestyle{empty}
\begin{abstract}
Scene graph generation has received growing attention with the advancements in image understanding tasks such as object detection, attributes and relationship prediction,~\etc. However, existing datasets are biased in terms of object and relationship labels, or often come with noisy and missing annotations, which makes the development of a reliable scene graph prediction model very challenging. In this paper, we propose a novel scene graph generation algorithm with external knowledge and image reconstruction loss to overcome these dataset issues. In particular, we extract commonsense knowledge from the external knowledge base to refine object and phrase features for improving generalizability in scene graph generation. To address the bias of noisy object annotations, we introduce an auxiliary image reconstruction path to regularize the scene graph generation network. Extensive experiments show that our framework can generate better scene graphs, achieving the state-of-the-art performance on two benchmark datasets: Visual Relationship Detection and Visual Genome datasets.
\end{abstract}
\section{Introduction}
With recent breakthroughs in deep learning and image recognition, higher-level visual understanding tasks, such as visual relationship detection, has been a popular research topic~\cite{elliott2013image,izadinia2014incorporating,gu2015recent,xiong2015recognize,yang2018shuffle}.
Scene graph, as an abstraction of objects and their complex relationships, {provides} {rich} semantic information of an image.
It involves the detection of {all} $\langle$\textit{subject}-\textit{predicate}-\textit{object}$\rangle$ triplets in an image and the localization of {all} objects. Scene graph provides a structured representation of an image that can support {a wide range of high-level visual tasks}, including image captioning~\cite{gu2017stack,gu2017empirical,gu2018unpaired,yang2018cap}, visual question answering~\cite{wang2018fvqa,wu2018image,zhao2018semantically}, image retrieval~\cite{gu2017look,johnson2015image}, and image generation~\cite{johnson2018image}. 
{However, it is not easy to extract scene graphs from images}, since it involves not only detecting and localizing pairs of interacting objects but also recognizing their pairwise relationships.
Currently, there are two categories of approaches for scene graph generation. Both categories group object proposals into pairs and use the phrase features (features of their union area) for predicate inference.
The difference of the two categories lies in the different procedures. The first category detects the objects first and then recognizes the relationships between those objects~\cite{dai2017detecting,liao2017natural, lu2016visual}. The second category {jointly identifies} the objects and their relationships based on the object and relationship proposals~\cite{li2017vip,li2018factorizable,wang2018scene}.

\begin{figure}[t!]
	\centering
	\includegraphics[width=\linewidth]{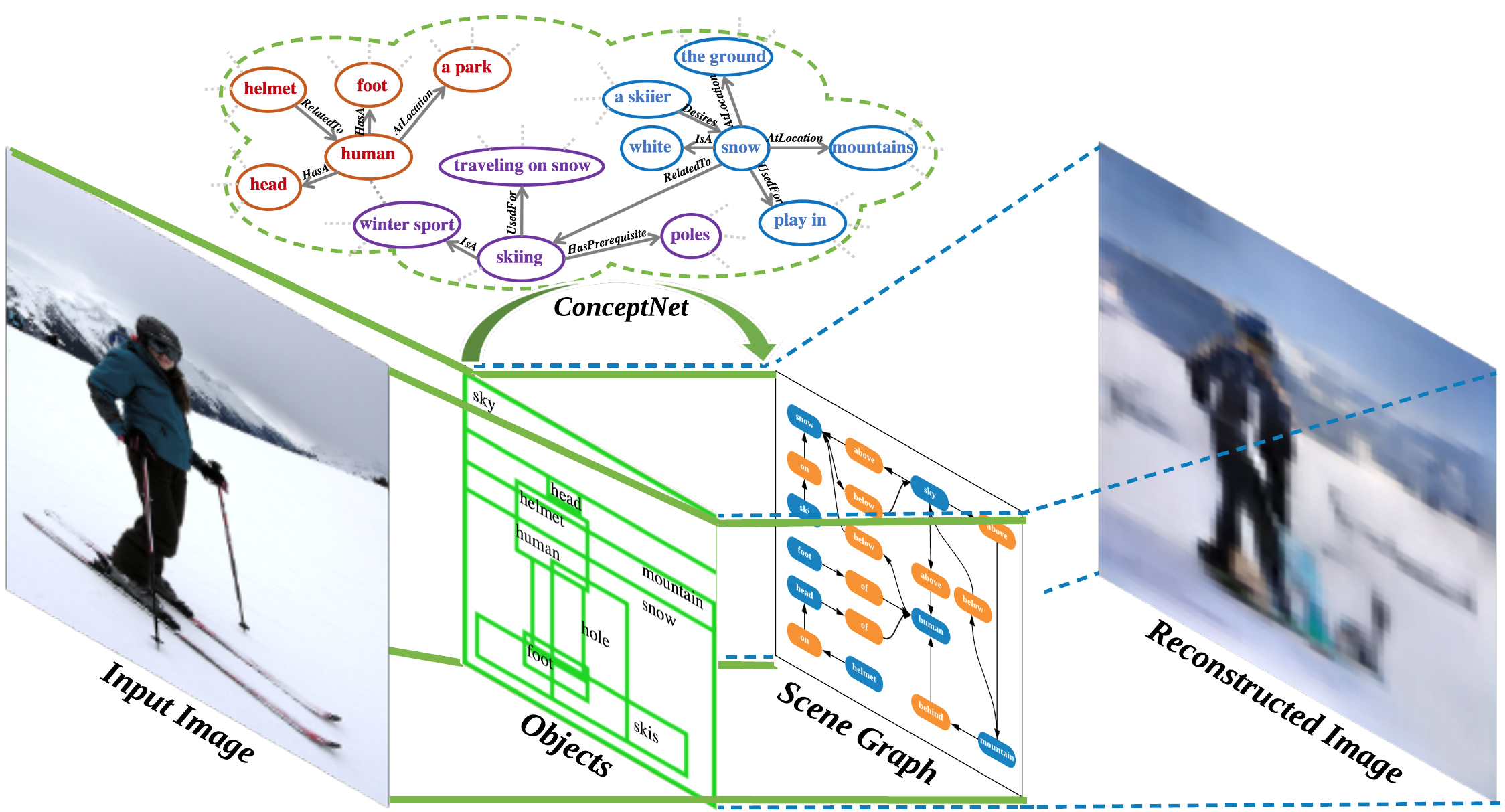}
	\vspace{-5mm}
	\caption{Conceptual illustration of our scene graph learning model. The left (\textit{green}) part illustrates the image to scene graph generation, {the right (\textit{blue}) part illustrates the image-level regularizer that reconstructs the image based on object labels and bounding boxes}. The commonsense knowledge reasoning (\textit{top}) is introduced to the scene graph generation process.}
	\label{fig:principle}
	\vspace{-5mm}
\end{figure}

{Despite the promising progress introduced by these approaches, most of them suffer from the limitations of existing scene graph datasets.}
First, to comprehensively depict an image using the scene graph, it requires a wide variety of relation triplets $\langle$\textit{subject}-\textit{predicate}-\textit{object}$\rangle$. Unfortunately, current datasets only capture a small portion of the knowledge~\cite{lu2016visual}, \eg, Visual Relationship Detection (VRD) dataset. Training on {such a dataset with long-tail distributions} will cause the prediction model bias towards those most-frequent relationships. Second, predicate labels are highly determined by the identification of object pairs~\cite{zellers2018neural}. However, due to the difficulty of exhaustively labeling bounding boxes of all instances of each object, the current large-scale crowd-sourced datasets like Visual Genome (VG)~\cite{krishna2017visual} are contaminated by noises (\eg, missing annotations and meaningless proposals). Such a noisy dataset will inevitably result in a poor performance of the trained object detector~\cite{bansal2018zero}, which further hinders the performance of predicate detection.

For human beings, we are capable of reasoning over visual elements of an image based on our commonsense knowledge. For example, in Figure~\ref{fig:principle}, humans have the background knowledge: the subject (\textit{woman}) {appears / stands} on something; the object (\textit{snow}) {enhances} the evidence of the predicate (\textit{skiing}). Commonsense knowledge can also help {correct} object detection. For example, the specific external knowledge for \textit{skiing} benefits {inference of} the object~(\textit{snow}) as well. This motivates us to {leverage} commonsense knowledge {to help} scene graph generation.

Meanwhile, despite the crucial role of object labels for relationship prediction, existing datasets are very noisy due to the significant amount of missing object annotations. However, our goal is to obtain scene graphs with more complete scene representation. Motivated by this goal, we regularize our scene graph generation network by reconstructing the image from detected objects. Considering the case in Figure~\ref{fig:principle}, a method might recognize \textit{snow} as \textit{grass} by mistake. If we generate an image based on the falsely predicted scene graph, this minor error would be heavily penalized, even though most of the \textit{snow}'s relationships might be correctly identified.

The contributions of this paper are threefold. 1) We propose a knowledge-based feature refinement module to incorporate commonsense knowledge from an external knowledge base. Specifically, the module extracts useful information from ConceptNet~\cite{speer2013conceptnet} to refine object and phrase features before scene graph generation. We exploit Dynamic Memory Network (DMN)~\cite{kumar2016ask} to implement multi-hop reasoning over the retrieved facts and infer the most probable relations accordingly. 2) We introduce image-level supervision {module} by reconstructing the image to regularize our scene graph generation model. We view this auxiliary branch as a regularizer, which is only present during training. 3) We conduct extensive experiments on two benchmark datasets: VRD and VG datasets. Our empirical results demonstrate that our approach can significantly {improve} the state-of-the-art on scene graph generation.
\section{Related Works}
\noindent\textbf{Incorporating Knowledge in Neural Networks.}
There has been {growing} interest in improving data-driven models with external Knowledge Bases (KBs) in natural language processing~\cite{hinton2015distilling,bao2014knowledge} and computer vision communities~\cite{li2017incorporating,aditya2018explicit,deng2014large}. Large-scale structured KBs are constructed either by manual effort (\eg, Wikipedia, DBpedia~\cite{auer2007dbpedia}), or by automatic extraction from unstructured or semi-structured data (\eg, ConceptNet). One direction to improve the data-driven model is to distill external knowledge into Deep Neural Networks~\cite{xiong2016dynamic,yu2017visual,hu2016deep}. Wu~\etal~\cite{wu2018image} encode the mined knowledge from DBpedia~\cite{auer2007dbpedia} {into} a vector and {combine it} with visual features to {predict answers}. Instead of aggregating the textual vectors with average-pooling operation~\cite{wu2018image}, Li~\etal~\cite{li2017incorporating} distill the retrieved context-relevant external knowledge triplet through a DMN for open-domain visual question answering. Unlike~\cite{wu2018image,li2017incorporating}, Yu~\etal~\cite{yu2017visual} extract linguistic knowledge from training annotations and Wikipedia, and distill knowledge to regularize training and provide extra cues for inference.
{A teacher-student framework is adopted to minimize} the KL-divergence of the prediction distributions of teacher and student.

\noindent\textbf{Visual Relationship Detection.}
Visual relationship detection has been investigated by many works in the last decade~\cite{johnson2015image,ding2019semantic,ding2018context,plummer2017phrase}. Lu~\etal~\cite{lu2016visual} introduce generic {visual relationship detection} as a visual task, where {they detect} objects first, and then {recognize} predicates between object pairs. Recently, some works have explored the message passing for context propagation and feature refinement~\cite{xu2017scene,li2017vip}. {Xu~\etal~\cite{xu2017scene} construct the scene graph by refining the object and relationship features jointly with message passing}.
Dai~\etal~\cite{dai2017detecting} exploit the statistical dependencies between objects and their relationships and refine the posterior probabilities iteratively with a Conditional Random Field (CRF) network.
{More recently, Zeller~\etal~\cite{zellers2018neural} achieve a strong baseline by predicting relationships with frequency priors}.
{To deal with the large number of potential relations between objects, Yang~\etal~\cite{yang2018graph} propose a relation proposal network that prunes out uncorrelated object pairs, and captures the contextual information with an attentional graph convolutional network}.
{In~\cite{li2018factorizable}, they propose a clustering method which factorizes the full graph into subgraphs, where each subgraph is composed of several objects and a subset of their relationships.}

{Most related to our work} are the approaches proposed by Li~\etal~\cite{li2018factorizable} and Yu~\etal~\cite{yu2017visual}.
Unlike~\cite{li2018factorizable}, which {focuses} on the efficient scene graph {generation, our} approach addresses the long tail distribution of relationships by commonsense cues along with visual cues.
{Unlike~\cite{yu2017visual}, which leverages linguistic knowledge to regularize the network, our knowledge-based module improves the feature refining procedure by reasoning over a basket of commonsense knowledge retrieved from ConceptNet}.

\section{Methodology}
\begin{figure*}[t!]
	\centering
	\includegraphics[width=\textwidth]{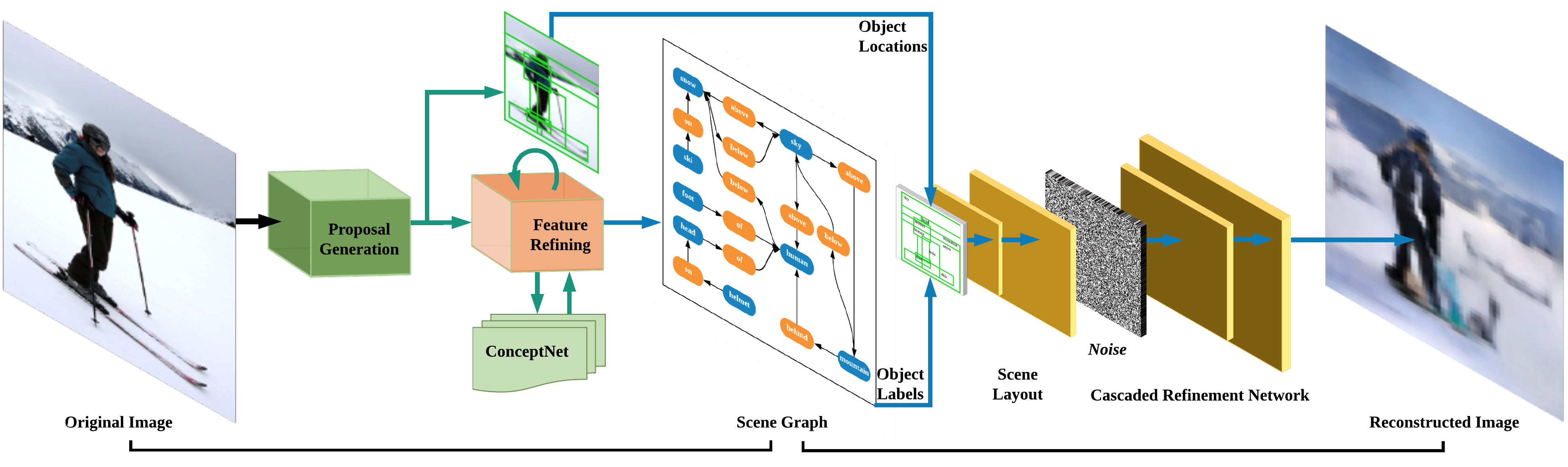}
	\vspace{-7mm}
	\caption{Overview of the proposed scene graph generation framework. {The left part generates a scene graph from the input image. The right part is an auxiliary image-level regularizer which reconstructs the image based on the detected object labels and bounding boxes. After training, we discard the image reconstruction branch.}}
	\label{fig:framework}
	\vspace{-5mm}
\end{figure*}

Figure~\ref{fig:framework} gives an overview of our proposed scene graph generation framework. The entire framework can be divided into the following steps:
(1) generate object and subgraph proposals for a given image;
(2) refine object and subgraph features with external knowledge;
(3) generate the scene graph by recognizing object categories with object features and recognizing object relations by fusing subgraph features and object feature pairs;
(4) reconstruct the {input} image via an additional generative path. During training, we use two types of supervisions: scene graph level supervision and image-level supervision. For scene graph level supervision, we optimize our model by {guiding} the generated scene graph with the ground truth {object and predicate categories}. The image-level supervision is introduced to overcome the aforementioned {missing annotations} by reconstructing the image from objects and {enforcing} the reconstructed image close to the original image.

\subsection{Proposal Generation}\label{sec:proposal}
\textbf{Object Proposal Generation.}
Given an image $\mathbf{I}$, we first use the Region Proposal Network (RPN)~\cite{ren2015faster} to extract a set of object proposals:
\begin{equation}
[\mathrm{o}_0,\cdots, \mathrm{o}_{N-1}] = f_{\text{RPN}}(\mathbf{I})
\end{equation}
where $f_{\text{RPN}}(\cdot)$ stands for the RPN module, and $o_i$ is the $i$-th object proposal represented by a bounding box $r_{i} =[x_i,y_i,w_i,h_i]$ with $(x_i,y_i)$ {being the coordinates of the top left corner and $w_i$ and $h_i$ being the width and the height of the bounding box, respectively}. For any two different objects $\langle o_i, o_j\rangle$, there are two possible relationships in opposite directions. Thus, for $N$ object proposals, there are totally $N(N-1)$ potential relations. Although more object proposals lead to a {bigger} scene graph, the number of potential relations will increase dramatically, which significantly increases {the} computational cost and deteriorates the inference speed. To address this issue, subgraph is introduced in~\cite{li2018factorizable} to reduce the number of potential relations by clustering. 

\textbf{Subgraph Proposal Construction.}
We adopt the {clustering} approach proposed in~\cite{li2018factorizable}. In particular, for a pair of object proposals, a subgraph proposal is constructed as the union box with the confidence score being the product of the scores of the two object proposals. Then, subgraph proposals are suppressed by non-maximum-suppression (NMS). In this way, a candidate relation can be represented by two objects and one subgraph: {$\langle{o}_i,{o}_j,{s}_k^i\rangle$}, where $i\neq j$ and $s_k^i$ is the $k$-th subgraph of all the subgraphs associated with {${o}_i$}, which contains {${o}_j$} as well as some other object proposals. Following~\cite{li2018factorizable}, we represent a subgraph and an object as a feature map, $\mathbf{s}_k^i\in \mathbb{R}^{D\times K_s \times K_s}$, and a feature vector, $\mathbf{o}_i\in \mathbb{R}^D$, respectively, where $D$ and $K_s$ are the dimensions.

\subsection{Feature Refinement with External Knowledge}\label{sec:feature_refine}
\textbf{Object and Subgraph Inter-refinement.} Considering that each object $\mathbf{o}_i$ is connected to a set of subgraphs $\mathbf{S}^i$ and each subgraph $\mathbf{s}_k$ is associated with a set of objects $\mathbf{O}^k$, we refine the object vector (resp. the subgraph) by attending the associated subgraph feature maps (resp. the associated object vectors):
\begin{align}
\bar{\mathbf{o}}_i=&\mathbf{o}_i + f_{s\rightarrow o}\left(\sum_{\mathbf{s}_k^i \in \mathbf{S}^i} \alpha_k^{s\rightarrow o} \cdot \mathbf{s}_k^i\right)\label{eq:message_pass_s2o_p}\\
\bar{\mathbf{s}}_k=&\mathbf{s}_k + f_{o\rightarrow s}\left(\sum_{\mathbf{o}_i^k \in \mathbf{O}^k} \alpha_i^{o\rightarrow s} \cdot \mathbf{o}_i^k\right)\label{eq:message_pass_o2s_p}
\end{align}
where $\alpha_k^{s\rightarrow o}$ (resp. $\alpha_i^{o\rightarrow s}$) is the output of a softmax layer indicating the weight for passing $\mathbf{s}_k^i$ (resp. $\mathbf{o}_i^k$) to $\mathbf{o}_i$ (resp. $\mathbf{s}_k$), and $f_{s\rightarrow o}$ and $f_{o\rightarrow s}$ are non-linear mapping functions. {This part is similar to~\cite{li2018factorizable}}. Note that due to different dimensions of $\mathbf{o}_i$ and $\mathbf{s}_k$, pooling or spatial location based attention needs to be respectively applied for $s\rightarrow o$ or $o \rightarrow s$ refinement. {Interested readers are referred to~\cite{li2018factorizable} for details}. 

\begin{figure}[t!]
	\centering
	\includegraphics[width=\linewidth]{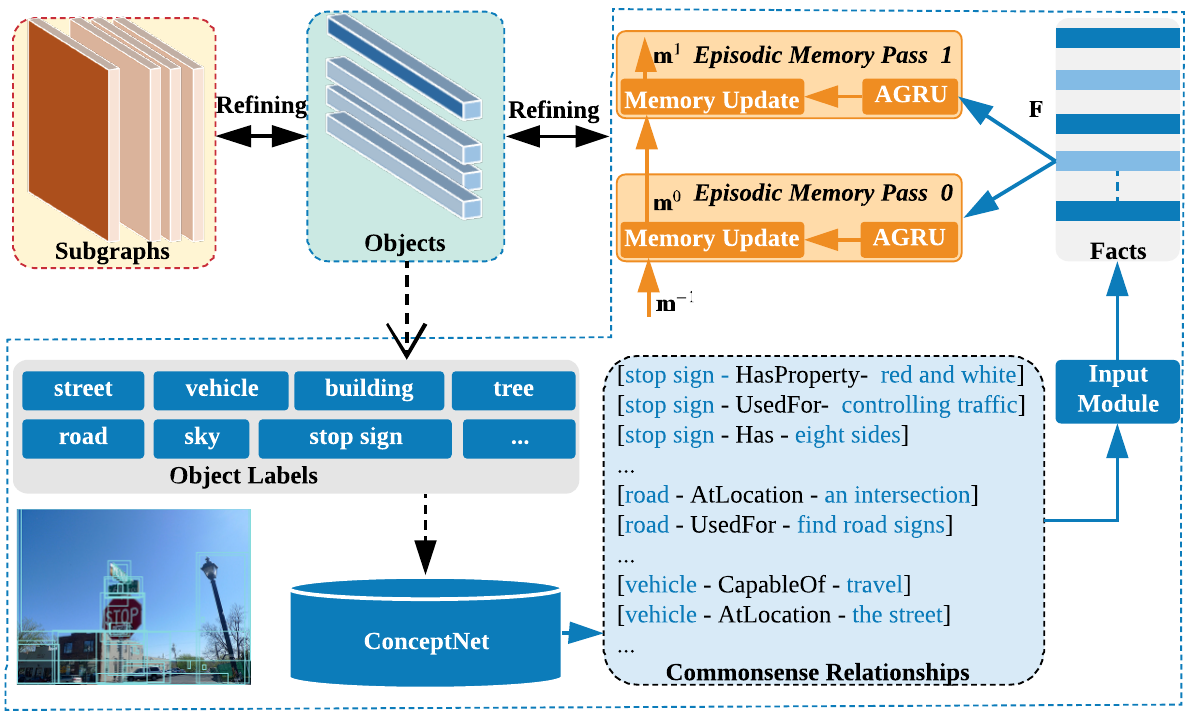}
	\vspace{-7mm}
	\caption{Illustration of our proposed knowledge-based feature refinement module. Given the object labels, we retrieve the facts (or symbolic triplets) from the ConceptNet (\textit{bottom}), and then reason those facts with dynamic memory network using two passes (\textit{top right}).}
	\label{fig:kdmn_v0}
	\vspace{-5mm}
\end{figure}

\textbf{Knowledge Retrieval and Embedding.}
To address the relationship distribution bias of the current visual relationship datasets, we propose a novel feature refinement network to further improve the feature representation by taking advantage of the commonsense relationships in external knowledge base (KB). In particular, we predict the object label $a_i$ from the refined object vector $\bar{\mathbf{o}}_i$, and match $a_i$ with the corresponding semantic entities in KB. Afterwards, we retrieve the corresponding commonsense relationships from KB using the object label $a_i$:
\begin{align}
{a}_i&\overset{\text{retrieve}}{\longrightarrow} \langle {a}_i,{a}_{i, j}^r, {a}_{j}^o, w_{i,j}\rangle, j\in [0,K-1]\label{eq:concept_net_triplet}
\end{align}
where $a^r_{i,j}$, $a^o_{j}$ and $w_{i,j}$ are the top-$K$ corresponding relationships, the object entity and the {weight, respectively}. Note that the weight $w_{i,j}$ is provided by KB (\ie, ConceptNet~\cite{speer2013conceptnet}), indicating how common a triplet $\langle {a}_i,{a}_{i,j}^r, {a}_{j}^o\rangle$ is. Based on the weight $w_{i,j}$, we can identify the top-$K$ most common relationships for $a_i$. Figure~\ref{fig:kdmn_v0} illustrates the process of our proposed knowledge-based feature refinement module.

To encode the retrieved commonsense relationships, we first transform each symbolic triplet $\langle {a}_i,{a}_{i,j}^r, {a}_{j}^o\rangle$ into a sequence of words: $[X^{0},\cdots, X^{T_a-1}]$, and then map each word in the sentence into a continuous vector space with word embedding $\mathbf{x}^{t}=\mathbf{W}_e X^{t}$.
The embedded vectors are then fed into an RNN-based encoder~\cite{xiong2016dynamic} as
\begin{equation}
\mathbf{h}_{k}^t = \textrm{RNN}_{\text{fact}}(\mathbf{x}_{k}^t, \mathbf{h}_{k}^{t-1}), \ t\in [0, T_a-1]\label{eq:pos_enc_rnn}
\end{equation}
where $\mathbf{x}_{k}^t$ is the $t$-th word embedding of the $k$-th sentence, and $\mathbf{h}_{k}^{t}$ is the hidden state of the encoder. We use a bi-directional Gated Recurrent Unit (GRU) for $\textrm{RNN}_{\text{fact}}$ and the final hidden state $\mathbf{h}_{k}^{T_a-1}$ is treated as the vector representation for the $k$-th retrieved sentence or fact, denoted as $\mathbf{f}_k^i$ for object $\mathbf{o}_i$.

\textbf{Attention-based Knowledge Fusion.}
The knowledge units are stored in memory slots for reasoning and updating.
{Our target is to incorporate the external knowledge into the procedure of feature refining.}
However, for $N$ objects, we have $N\times K$ relevant fact vectors in memory slots. This makes it difficult to distill the useful information from the candidate knowledge when $N\times K$ is large.
{DMN~\cite{kumar2016ask} provides a mechanism to pick out the most relevant facts by using an episodic memory module. Inspired by this, we adopt the improved DMN~\cite{xiong2016dynamic} to reason over the retrieved facts $\mathbf{F}$, where $\mathbf{F}$ denotes the set of fact embedding $\{\mathbf{f}_k\}$.}
It consists of an attention component which generates a contextual vector using the episode memory $\mathbf{m}^{t-1}$.
Specifically, we feed the object vector $\mathbf{\bar{o}}$ to a non-linear fully-connected layer and attend the facts as follows:
\begin{align}
\mathbf{q} =& \tanh (\mathbf{W}_q\mathbf{\bar{o}}+\mathbf{b}_q)\label{eq:dmn_question}\\
	\mathbf{z}^t =& [\mathbf{F}\circ \mathbf{q};\mathbf{F}\circ \mathbf{m}^{t-1};|\mathbf{F}-\mathbf{q}|;|\mathbf{F}-\mathbf{m}^{t-1}|]\\
	\mathbf{g}^t=&\text{softmax}(\mathbf{W}_1\tanh (\mathbf{W}_2\mathbf{z}^t+\mathbf{b}_2)+\mathbf{b}_1)\label{eq:dmn_attention}\\
	\mathbf{e}^t=&\text{AGRU}(\mathbf{F}, \mathbf{g}^t) \label{eq:fet}
\end{align}
where $\mathbf{z}^t$ is the interactions between the facts $\mathbf{F}$, the episode memory $\mathbf{m}^{t-1}$ and the mapped object vector $\mathbf{q}$, $\mathbf{g}^t$ is the output of a softmax layer, $\circ$ is the element-wise product, $|\cdot|$ is the element-wise absolute value, and $[\ ; \ ]$ is the concatenation operation. Note that $\mathbf{q}$ and $\mathbf{m}$ need to be expanded via duplication in order to have the same dimension as $\mathbf{F}$ for the interactions. In~\eqref{eq:fet}, $\text{AGRU}(\cdot)$ refers to the Attention based GRU~\cite{xiong2016dynamic} which replaces the update gate in GRU with the output attention weight $\mathbf{g}_k^t$ for fact $k$: 
\begin{align}
	\mathbf{e}_k^t =& g_k^t \text{GRU}(\mathbf{f}_k,\mathbf{e}_{k-1}^t)+(1-g_k^t)\mathbf{e}_{k-1}^t
\end{align}
where $\mathbf{e}_{K}^t$ is the final state of the episode which is the state of the GRU after all the $K$ sentences have been seen.

After one pass of the attention mechanism, the memory is updated using the current episode state and the previous memory state:
\begin{align}
		\mathbf{m}^t = \text{ReLU}(\mathbf{W}_m[\mathbf{m}^{t-1};\mathbf{e}_K^t;\mathbf{q}]+\mathbf{b}_m)\label{eq:dmn_memory} .
\end{align}
where $\mathbf{m}^t$ is the new episode memory state. {By the final pass $T_m$, the episodic memory $\mathbf{m}^{T_m-1}$ can memorizes useful knowledge information for relationship prediction.}

{The final episodic memory $\mathbf{m}^{T_m-1}$ is passed to refine the object feature $\mathbf{\bar{o}}$ as}
\begin{align}
\mathbf{\tilde{o}} =\text{ReLU}(\mathbf{W}_c[\mathbf{\bar{o}};\mathbf{m}^{T_m-1}]+\mathbf{b}_c)\label{eq:dmn_refine}
\end{align}
where $\mathbf{W}_c$ and $\mathbf{b}_c$ are parameters to be learned.
{In particular, we refine objects with KB via \eqref{eq:dmn_refine} as well as jointly refining objects and subgraphs by replacing $\{\mathbf{{o}}_i,\mathbf{{s}}_i\}$ with $\{\mathbf{\tilde{o}}_i,\mathbf{\bar{s}}_i\}$ in \eqref{eq:message_pass_s2o_p} and \eqref{eq:message_pass_o2s_p}, in an iterative fashion (see Alg.~\ref{alg_full})}.

\subsection{Scene Graph Generation}\label{sec:inference}
\textbf{Relation Prediction.}
After the feature refinement, we can predict object labels as well as predicate labels with the refined object and subgraph features.
For object label, we can predict it directly with the object features.
{For relationship label, as the subgraph feature is related to several object pairs, we predict the label based on subject and object feature vectors along with their corresponding subgraph feature map. We formulate the inference process as}
\begin{align}
\mathrm{P}_{i,j}\sim & \text{softmax}(f_{\text{rel}}([\mathbf{\tilde{o}}_i\otimes \mathbf{\bar{s}}_k; \mathbf{\tilde{o}}_j\otimes \mathbf{\bar{s}}_k;\mathbf{\bar{s}}_k]))\\
\mathrm{V}_i\sim& \text{softmax}(f_{\text{node}}(\mathbf{\tilde{o}}_i))
\end{align}
where $f_{\text{rel}}(\cdot)$ and $f_{\text{node}}(\cdot)$ denote the mapping layers for predicate and object recognition, respectively, and $\otimes$ denotes the convolution operation~\cite{li2018factorizable}. Then, we can construct the scene graph as: $\mathcal{G}=\langle V_i, P_{i,j}, V_j\rangle, i\neq j$.

\textbf{Scene Graph Level Supervision.}
Like other approaches~\cite{li2017msdn, li2018factorizable, wang2018scene}, during training we want the generated scene graph close to the ground-truth scene graph by optimizing the scene graph generation process with object detection loss and relationship classification loss
\begin{equation}
\mathcal{L}_{\text{im2sg}}=\lambda_{\text{pred}}\mathcal{L}_{\text{pred}}+\lambda_{\text{obj}}\mathcal{L}_{\text{obj}}+\lambda_{\text{reg}} \mathbf{1}_{ u \ge 1} \mathcal{L}_{\text{reg}}\label{eq:im2sg_final_loss}
\end{equation}
where $\mathcal{L}_{\text{pred}}$, $\mathcal{L}_{\text{obj}}$ and $\mathcal{L}_{\text{reg}}$ are the predicate classification loss, the object classification loss and the bounding box regression loss, respectively, $\lambda_{\text{obj}}$, $\lambda_{\text{pred}}$ and $\lambda_{\text{reg}}$ are hyper-parameters, and $\mathbf{1}$ is the indicator function with $u$ being the object label, $u\ge 1$ for object categories and $u=0$ for background.

For the predicate detection, the output is the probability over all the candidate predicates. $\mathcal{L}_{\text{pred}}$ is defined as the softmax loss. Like the predicate classification, the output of the object detection is the probability over all the object categories. $\mathcal{L}_{\text{cls}}$ is also defined as the softmax loss. For the bounding box regression loss $\mathcal{L}_{\text{reg}}$, we use smooth $ L_1 $ loss~\cite{ren2015faster}. 

\begin{figure}[t!]
	\centering
	\includegraphics[width=\linewidth]{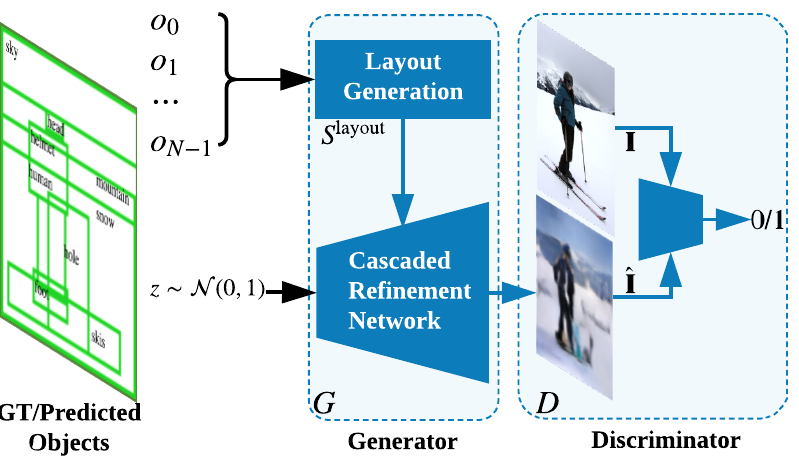}
	\vspace{-7mm}
	\caption{Illustration of our proposed object-to-image generation module $\text{Gen}_{\text{o2i}}$.}
	\label{fig:sg2im_v0}
	\vspace{-5mm}
\end{figure}

\subsection{Image Generation}
To better regularize the networks, an object-to-image generative path is added. Figure~\ref{fig:sg2im_v0} depicts our proposed object-to-image generation module $\text{Gen}_{\text{o2i}}$.
{In particular, we first compute a scene layout based on the object labels and their corresponding locations. For each object $i$, we expand the object embedding vectors $\mathbf{o}_i\in \mathbb{R}^D$ to shape $D\times 8 \times 8$, and then wrap it to the position of the bounding box $r_i$ using bilinear interpolation to give an object layout $o_i^{\text{layout}}\in \mathbb{R}^{D\times H\times W}$, where $D$ is the dimension of the embedding vectors for objects and $H\times W=64\times 64$ is the output image resolution. We sum all object layouts to obtain the scene layout $S^{\text{layout}}=\sum_i o_i^{\text{layout}}$.}

Given the scene layout, we synthesize an image that respects the object positions with an image generator $G$. Here, we adopt a cascaded refinement network~\cite{johnson2018image} which consists of a series of convolutional refinement modules to generate the image. The spatial resolution doubles between the convolutional refinement modules. This allows the generation to proceed in a coarse-to-fine manner.
For each module, it takes two inputs. One is the output from the previous module (the first module takes Gaussian noise), and the other one is the scene layout $S^{\text{layout}}$, which is downsampled to the input resolution of the module. 
These inputs are concatenated channel-wisely and passed to a pair of $3\times 3$ convolution layers. The outputs are then upsampled using nearest-neighbor interpolation before being passed to the next module.
The output from the last module is passed to two final convolution layers to produce the output image.

\textbf{Image-level Supervision.} In addition to the common pixel reconstruction loss $\mathcal{L}_{\text{pixel}}$, we also adopt a conditional GAN loss~\cite{reed2016generative}, considering the image is generated based on the objects. In particular, we train the discriminator $D_i$ and the generator $G_i$ by alternatively maximizing $\mathcal{L}_{D_i}$ in Eq.~\eqref{eq:gan_d_loss} and $\mathcal{L}_{G_i}$ in Eq.~\eqref{eq:gan_g_loss}:
\begin{align}
\mathcal{L}_{D_i}=&\mathbb{E}_{I\sim p_{\text{real}}}[\log D_i(\mathbf{I})]\label{eq:gan_d_loss}\\
\mathcal{L}_{G_i}=&\mathbb{E}_{\hat{I}\sim p_{\text{G}}}[\log (1-D_i(\hat{\mathbf{I}})]+ \lambda_{p}\mathcal{L}_{\text{pixel}}\label{eq:gan_g_loss}
\end{align}
where $\lambda_{p}$ is the tuning parameter. 
For the generator loss, we maximize $\log D_i(G_i(z|S^{\text{layout}}))$ rather than minimizing the original $\log(1-D_i(G_i(z|S^{\text{layout}})))$ for better gradient behavior.
For the pixel reconstruction loss, we calculate the $\ell_1$ distance between the real image $\mathbf{I}$ and a corresponding synthetic image $\hat{\mathbf{I}}$ as $||\mathbf{I}-\hat{\mathbf{I}}||_1$.

{As shown in Figure~\ref{fig:framework}, we view the object-to-image generation branch as a regularizer. It can be seen as a corrective model for scene graph generation by improving the performance of object detection. During training, backpropagation from losses \eqref{eq:im2sg_final_loss},~\eqref{eq:gan_d_loss}, and~\eqref{eq:gan_g_loss} influences the model parameter updates.}
This image-level supervision can be seen as a corrective model for scene graph generation by improving the performance of object detection. The gradients back-propagated from the object-to-image branch update the parameters of our object detector and the feature refinement module which is followed by the relation prediction.

Alg.~\ref{alg_full} summarizes the entire training procedure.
\begin{algorithm}
	\caption{Training procedure.}
	\label{alg_full}
	\begin{algorithmic}[1]
		\Require{Image $\mathbf{I}$, number of training steps $T_s$.}
		\State Pretrain image generation module $\text{Gen}_{\text{o2i}}$ (GT objects)
		\For{$t=0:T_m-1$}
		\State Get objects and relationship triples.
		\State Proposal Generation: $(\mathbf{O},\mathbf{S}) \leftarrow \mathbf{I}$ \{RPN\}
		\State{/*{Knowledge-based Feature Refining}*/}
		\For{$r=0:T_r-1$}
		\State $\bar{\mathbf{o}}_i \leftarrow \{\mathbf{o}_i, \mathbf{S}^i\}$ /*Refining using \eqref{eq:message_pass_s2o_p}*/
		\State $\bar{\mathbf{s}}_k \leftarrow \{\mathbf{s}_k, \mathbf{O}^k\}$ /*Refining using \eqref{eq:message_pass_o2s_p}*/
		\State $\tilde{\mathbf{o}}_i \leftarrow \{\mathbf{F}, \mathbf{\bar{o}_i}\}$ /*Refining using \eqref{eq:dmn_refine}*/
		\State ${\mathbf{o}}_i \leftarrow \tilde{\mathbf{o}}_i$, ${\mathbf{s}}_i \leftarrow \bar{\mathbf{s}}_i$
		\EndFor
		\State Update parameters with $\text{Gen}_{\text{o2i}}$ (predicted objects)
		\State Update parameters with \eqref{eq:im2sg_final_loss}
		\EndFor
		
		\setcounter{ALG@line}{0}
		\Statex{\hspace{-18pt}\textbf{Function: }}{$\text{Gen}_{\text{o2i}}$}
		\Require{Real image $\mathbf{I}$, objects (GT / predicted).}
		\State Object Layout Generation: $o_i^{\text{layout}}\leftarrow \{\mathbf{o}_i,r_i\}$
		\State Scene Layout Generation: $S^{\text{layout}}=\sum_i o_i^{\text{layout}}$
		\State Image Reconstruction: $\hat{\mathbf{I}}=G_i(z, S^{\text{layout}})$
		\State Update image generator $G_i$ parameters using \eqref{eq:gan_g_loss}.
		\State Update image discriminator $D_i$ parameters using \eqref{eq:gan_d_loss}.
	\end{algorithmic}
	\label{ag:training_process}
\end{algorithm}
\vspace{-6mm}
\section{Experiments}
\subsection{Datasets}
We evaluate our approach on two datasets: VRD~\cite{lu2016visual} and VG~\cite{li2017msdn}.
VRD is the most widely used benchmark dataset for {visual relationship detection}. Compared with VRD, the raw VG~\cite{krishna2017visual} contains a large number of noisy labels. In our experiment, we use a cleansed-version VG-MSDN in~\cite{li2017msdn}. Detailed statistics of both datasets are shown in Table~\ref{tab:dataset}.

For the external KB, we employ the English subgraph of ConceptNet~\cite{speer2013conceptnet} as our knowledge graph.
{ConceptNet is a large-scale graph of general knowledge which aims to align its knowledge resources on its core set of 40 relations}. A large portion of these relation types can be considered as visual relations, {such as}, spatial co-occurrence (\eg, \textit{AtLocation}, \textit{LocatedNear}), visual properties of objects (\eg, \textit{HasProperty}, \textit{PartOf}), and actions (\eg, \textit{CapableOf}, \textit{UsedFor}).

\begin{table}[ht]
	\renewcommand{\arraystretch}{1}
	\setlength{\tabcolsep}{2pt}
	\small
	\caption{Dataset statistics. \#{Img} and {\#Rel}  denote the number of images and relation pairs respectively, {\#{Obj}} denotes the number of object categories, and \#{Pred} denotes the number of predicate categories.}
	\vspace{-5mm}
	\begin{center}
		\begin{tabularx}{1.0\linewidth}{l|ll|ll|l|l }
			\hline
			\multirow{2}{*}{\textbf{Dataset}} & \multicolumn{2}{c}{\textbf{Training Set}} \vline & \multicolumn{2}{c}{\textbf{Testing Set}}\vline & \multirow{2}{*}{\#Obj} & \multirow{2}{*}{\#{Pred}} \\
			& \#{Img} & \#{Rel} & \#{Img} & \#Rel & \\\hline
			VRD~\cite{lu2016visual} & 4,000 & 30,355 & 1,000 & 7,638 & 100 & 70 \\
			VG-MSDN~\cite{li2017msdn} & 46,164 & 507,296 & 10,000 & 111,396 & 150 & 50 \\
			\hline
		\end{tabularx}
	\end{center}
	\vspace{-5mm}
	\label{tab:dataset}
\end{table}
\subsection{Implementation Details}
As shown in Alg.~\ref{ag:training_process}, {we train our model in two phrases}. The initial phase looks only at the object annotations of the training set, ignoring the relationship triplets. For each dataset, we filter the objects according to the category and relation vocabularies in Table~\ref{tab:dataset}. We {then} learn an image-level regularizer that reconstructs the image based on the object labels and bounding boxes. The output size of the image generator is $64\times 64\times 3$, and the real image is resized before inputting to the discriminator. We train the regularizer with learning rate $10^{-4}$ and batch size 32. For each mini-batch we first update $G_i$, and then update $D_i$.

The second phase jointly trains the scene graph generation model and the auxiliary reconstruction branch. We adopt the Faster R-CNN~\cite{ren2015faster} associated with VGG-16~\cite{simonyan2014very} as the backbone.
{During training, the number of object proposals is 256.}
For each proposal, we use ROI align~\cite{he2017mask} pooling to generate object and subgraph features. The subgraph regions are pooled to $5\times 5$ feature maps. The dimension $D$ of the pooled object vector and the subgraph feature map is set to 512. For the knowledge-based refinement module, we set the dimension of word embedding to 300 and initialize it with the GloVe 6B pre-trained word vectors~\cite{pennington2014glove}.
{We keep the top-8 commonsense relationships.}
The number of hidden units of the fact encoder is set to 300, and the dimension of episodic memory is set to 512. {The iteration number $T_m$ of DMN update is set to 2.}
For the relation inference module, we adopt the same bottleneck layer as~\cite{li2018factorizable}. All the newly introduced layers are randomly initialized except the auxiliary regularizer. We set $\lambda_{\text{pred}}=2.0$, $\lambda_{\text{cls}}=1.0$, and $\lambda_{\text{reg}}=0.5$ in Eq~\eqref{eq:im2sg_final_loss}. The hyperparameter $\lambda_{p}$ in Eq~\eqref{eq:gan_g_loss} is set to 1.0. {The iteration number $T_r$ of the feature refinement is set to 2. We first train RPNs and then jointly train the entire network. The initial learning rate is 0.01, decay rate is 0.1, and stochastic gradient descent (SGD) is used as the optimizer. We deploy weight decay and dropout to prevent over-fitting.}

During testing, the image reconstruction branch will be discarded. {We respectively set the RPN non-maximum suppression (NMS)~\cite{ren2015faster} threshold to 0.6 and subgraph clustering~\cite{li2018factorizable} threshold to 0.5}. We output all the predicates and use the top-1 category as the prediction for objects and relations. Models are evaluated on two tasks: {Visual Phrase Detection (\textbf{PhrDet})} and {Scene Graph Generation (\textbf{SGGen})}.
{\textbf{PhrDet}} is to detect the $\langle$subject-predicate-object$\rangle$ phrases. {\textbf{SGGen}} is to detect the objects within the image and recognize their pairwise relationships.
Following~\cite{lu2016visual,li2018factorizable},  the Top-$K$ Recall~(denoted as Rec@$K$) is used as the performance metric; it calculates how many labeled relationships are hit in the top K predictions. In our experiments, Rec@50 and Rec@100 are reported.
Note that, Li~\etal~\cite{li2017msdn} and Yang~\etal~\cite{yang2018graph} reported the results on two more metrics: \textit{Predicate Recognition} and \textit{Phrase Recognition}. These two evaluation metrics are based on ground-truth object locations, which is not the case we consider. In our setting, we use detected objects for image reconstruction and scene graph generation. To be consistent with the training, we choose \textit{PhrDet} and \textit{SGGen} as the evaluation metrics, which is also more practical.

\subsection{Baseline Approaches for Comparisons}
\noindent\textbf{Baseline.}
This baseline model is the re-implementation of Factorizable Net~\cite{li2018factorizable}. We re-train it based on our backbone. Specifically, we use the same RPN model, and jointly train the scene graph generator until convergence.

\noindent\textbf{KB.}
This model is a KB-enhanced version of the baseline model. External knowledge triples are incorporated in DMN. The explicit knowledge-based reasoning is incorporated in the feature refining procedure.

\noindent\textbf{GAN.}
This model improves the baseline model by attaching an auxiliary branch that generates the image from objects with GAN. We train this model in two phases. The first phase trains the image reconstruction branch only with the object annotations. Then we refine the model jointly with the scene graph generation model.

\noindent\textbf{KB-GAN.}
This is our full model containing both KB and GAN. It is initialized with the trained parameters from KB and GAN, and fine-tuned with Alg.~\ref{ag:training_process}.

\subsection{Quantitative Results}
In this section, we present our quantitative results and analysis. To verify the effectiveness of our approach and analyze the contribution of each component, we first compare different baselines in Table~\ref{tab:ablaiton_study}, and investigate the improvement in recognizing objects in Table~\ref{tab:object_detection}. Then, we conduct a simulation experiment on VRD to investigate the effectiveness of our auxiliary regularizer in Table~\ref{tab:spare_data}. The comparison of our approach with the state-of-the-art methods is reported in Table~\ref{tab:comparison}.

\noindent\textbf{Component Analysis.}
In our framework, we proposed two novel modules -- KB-based feature refinement (KB) and auxiliary image generation (GAN). To get a clear sense of how these components affect the final performance, we perform ablation studies in Table~\ref{tab:ablaiton_study}. The left-most columns in Table~\ref{tab:ablaiton_study} indicate whether or not we use KB and GAN in our approach.
{To further investigate the improvement of our approach on recognizing objects, we also report object detection performance mAP~\cite{everingham2010pascal} in Table~\ref{tab:object_detection}}.

\begin{table}[ht]
\renewcommand{\arraystretch}{1}
\setlength{\tabcolsep}{3pt}
\small
\caption{Ablation studies of individual components of our method on VRD.}
\vspace{-6mm}
\begin{center}
\begin{tabularx}{1.0\linewidth}{l|l|ss|ss }
\hline
\multirow{2}{*}{\textbf{KB}} & \multirow{2}{*}{\textbf{GAN}} & \multicolumn{2}{c|}{\textbf{PhrDet}} & \multicolumn{2}{c}{\textbf{SGGen}}\\
& & Rec@50& Rec@100 & Rec@50& Rec@100 \\
\hline
- &- &25.57 & 31.09 &18.16 & 22.30\\
\checkmark& - & {27.02} & {34.04} &{19.85} & {24.58}\\
- & \checkmark  &26.65 & 34.06  & 19.56 & 24.64\\
\checkmark& \checkmark & \textbf{27.39} & \textbf{34.38} &\textbf{20.31} & \textbf{25.01}\\
\hline
\end{tabularx}
\end{center}\label{tab:ablaiton_study}
\vspace{-6mm}
\end{table}
\begin{table}[ht]
	\renewcommand{\arraystretch}{1.1}
	\setlength{\tabcolsep}{2.5pt}
	\small
    \caption{ Ablation study of the object detection on VRD.}
	\vspace{-5mm}
	\begin{center}
	\small
		\begin{tabularx}{\linewidth}{l|l|l|l|l|l|l}
			\hline
			\textbf{Model} &  \begin{tabular}{@{}l@{}}{Faster} \\ {R-CNN}~\cite{ren2015faster}\end{tabular} & \begin{tabular}{@{}l@{}}{ViP-}\ \\ {CNN}~\cite{li2017vip}\end{tabular}&{Baseline} & {KB} & {GAN} & \begin{tabular}{@{}l@{}}{KB-} \\ {GAN}\end{tabular} \\
			\hline
			mAP & 14.35 & 20.56 & 20.70 & 22.26 & 22.10  &\textbf{22.49} \\
			\hline
		\end{tabularx}
	\end{center}
	\vspace{-5mm}
	\label{tab:object_detection}
\end{table}
In Table~\ref{tab:ablaiton_study}, we observe that KB boosts \textbf{PhrDet} and \textbf{SGGen} significantly. This indicates our knowledge-based feature refinement can effectively learn the commonsense knowledge of objects to achieve high recall for the correct relationships. By adding the image-level supervision to the baseline model, the performance is further improved. This improvement demonstrates that the proposed image-level supervision is capable of capturing meaningful context across the objects. These results align with our intuitions discussed in the introduction. With KB and GAN, our model can generate scene graphs with high recall.

Table~\ref{tab:object_detection} demonstrates the improvement in recognizing objects. We can see that our full model (KB-GAN) outperforms Faster R-CNN~\cite{ren2015faster}, ViP-CNN~\cite{li2017vip} measured by mAP. It is worth noticing that the huge gain of KB illustrates that the introduction of commonsense knowledge substantially contributes to the object detection task.

\begin{table}[ht]
	\renewcommand{\arraystretch}{1}
	\setlength{\tabcolsep}{3pt}
	\small
	\caption{Ablation study of image-level supervision on subsampled VRD.}
    \vspace{-6mm}
	\begin{center}
		\begin{tabularx}{1.0\linewidth}{l|l|ss|ss }
			\hline
			 \multirow{2}{*}{\textbf{KB}} & \multirow{2}{*}{\textbf{GAN}} & \multicolumn{2}{c|}{\textbf{PhrDet}} & \multicolumn{2}{c}{\textbf{SGGen}}\\
			& & Rec@50& Rec@100 & Rec@50& Rec@100 \\
			\hline
			- &- & 15.44 & 20.96  & 10.94 & 14.53 \\
			- & \checkmark & 24.07 & 30.89  & 17.50 & 22.31  \\
			\checkmark& \checkmark & \textbf{26.62} & \textbf{31.13} &\textbf{19.78} & \textbf{24.17}\\
			\hline
		\end{tabularx}
	\end{center}
\vspace{-6mm}
	\label{tab:spare_data}
\end{table}
\begin{table*}[!htb]
	\renewcommand{\arraystretch}{1.1}
	\caption{Comparison with existing methods on \textbf{PhrDet} and \textbf{SGGen}.}
	\vspace{-6mm}
	\begin{center}
		\begin{tabularx}{1.0\linewidth}{l | l | ss| ss  }
		\hline
		\multirow{2}{*}{\textbf{Dataset}} & \multirow{2}{*}{\textbf{Model}}  & \multicolumn{2}{c}{\textbf{PhrDet}} \vline & \multicolumn{2}{c}{\textbf{SGGen}}  \\
		&&  Rec@50 & Rec@100 & Rec@50 & Rec@100 \\
		\hline
		\multirow{5}{*}{VRD~\cite{lu2016visual}} 
		& ViP-CNN~\cite{li2017vip} & 22.78 & 27.91 & 17.32 & 20.01 \\
		& DR-Net~\cite{dai2017detecting} & 19.93 & 23.45 & 17.73 & 20.88 \\
		& U+W+SF+LK: T+S~\cite{yu2017visual} &  {26.32} & {29.43} & {19.17} & {21.34} \\
		& Factorizable Net~\cite{li2018factorizable} &  {26.03} & {30.77} & {18.32} & {21.20} \\
		\cline{2-6}
		& \textbf{KB-GAN} &\textbf{27.39} & \textbf{34.38} &\textbf{20.31} & \textbf{25.01}\\
		\hline
		\hline
		\multirow{3}{*}{VG-MSDN~\cite{li2017msdn}} 
		& ISGG~\cite{xu2017scene} &  15.87 & 19.45 & 8.23  &   10.88\\
		& MSDN~\cite{li2017msdn} & 19.95 & 24.93 & 10.72 & 14.22  \\
        & Graph R-CNN~\cite{yang2018graph} &  -- & -- & {11.40} & {13.70} \\
		& Factorizable Net~\cite{li2018factorizable} & {22.84}  & {28.57} & {13.06} & {16.47}\\
		\cline{2-6}
		&\textbf{KB-GAN} &\textbf{23.51} &\textbf{30.04} &\textbf{13.65} & \textbf{17.57}\\
		\hline	
		\end{tabularx}
	\end{center}
    \vspace{-6mm}
	\label{tab:comparison}
\end{table*}
\begin{figure*}[ht!]
	\centering
	\includegraphics[width=\textwidth]{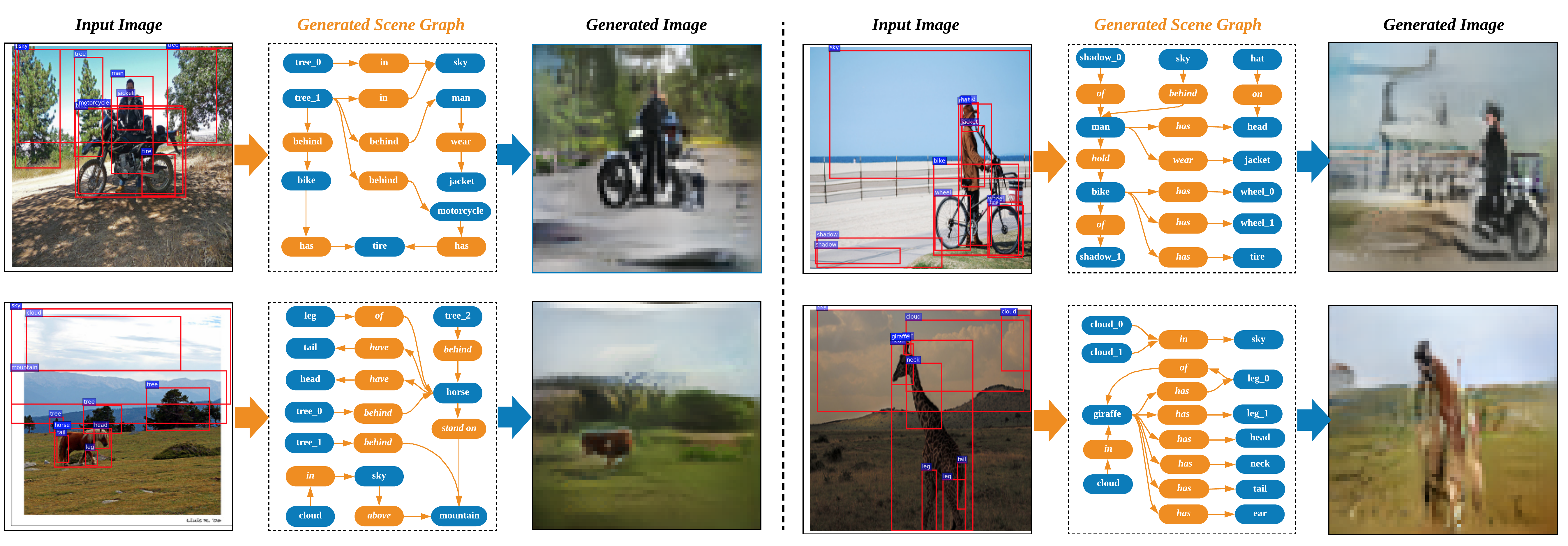}
	\vspace{-5mm}
	\caption{Qualitative results from KB-GAN. In each example, the left image is the original input image; the scene graph is generated by KB-GAN; and the right image is reconstructed from the detected objects.}
	\vspace{-5mm}
	\label{fig:eg1}
\end{figure*}

\noindent\textbf{Investigation on Image-level Supervision.}
As aforementioned, our image-level supervision can exploit the instances of rare categories. To demonstrate that our introduced image-level supervision can help on this issue, we exaggerate the problem by randomly removing 20\% object instances as well as their corresponding relationships from the dataset.
In Table~\ref{tab:spare_data}, we can see that training on such a sub-sampled dataset (with only 80\% object instances), Rec@50 of the baseline model drops from 25.57 (resp. 18.16) to 15.44 (resp. 10.94) for PhrDet and SGGen. However, with the help of GAN, Rec@50 of our final model decreases only slightly from 27.39 (resp. 20.31) to 26.62 (resp. 19.78) for PhrDet and SGGen, respectively.

We give our explanation on this significant performance improvement as below. Too many low-frequency categories deteriorate the training gain when only utilizing the class label as training targets. With the explicit image-level supervision, the proposed image reconstruction path can utilize the large quantities of instances of rare classes. This image-level supervision idea is generic, which can apply to many potential applications such as object detection.

\noindent\textbf{Comparison with Existing Methods.}
Table~\ref{tab:comparison} shows the comparison of our approach with the existing methods. We can see that our proposed method outperforms all the existing methods in the recall on both datasets. Compared with these methods, our model recognizes the objects and their relationships not only in the graph domain but also in the image domain.

\subsection{Qualitative Results}
Figure~\ref{fig:eg1} visualizes some examples of our full-model. We show the generated scene graph as well as the reconstructed image for each sample. It is clear that our method can generate high-quality relationship predictions in the generated scene graph. Also notable is that our auxiliary output images are reasonable. This demonstrates our model's capability to generate rich scene graph by learning with both external KB and auxiliary image-level regularizer.

\section{Conclusion}
In this work, we have introduced a new model for scene graph generation which includes a novel knowledge-base feature refinement network that effectively propagates contextual information across the graph, and an image-level supervision that regularizes the scene graph generation from image domain. Our framework outperforms state-of-the-art methods for scene graph generation on VRD and VG datasets. 
Our experiments show that it is fruitful to incorporate the commonsense knowledge as well as the image-level supervision into the scene graph generation. Our work shows a promising way to improve high-level image understanding via scene graph.

\section*{Acknowledgments}
This work was supported in part by Adobe Research, NTU-IGS, NTU-Alibaba Lab, and NTU ROSE Lab.

\newpage
\bibliographystyle{ieee}
\bibliography{egbib}

\end{document}